\documentclass{article}

\usepackage{microtype}
\usepackage{graphicx}
\usepackage{subfigure}
\usepackage{booktabs} 
\usepackage{natbib}
\usepackage{amsmath}
\usepackage{amsthm}
\usepackage{amsfonts}

\newtheorem{lemma}{Lemma}
\newtheorem{theorem}{Theorem}

\usepackage{hhline}
\usepackage{multirow}
\usepackage{hyperref}
\usepackage{ulem}
\usepackage[accepted]{icml2021}

\newcommand{\E}{\mathbb{E}}

\icmltitlerunning{Accurate and Reliable Forecasting using Stochastic Differential Equations}

\begin{document}
\twocolumn[
\icmltitle{
Accurate and Reliable Forecasting using Stochastic Differential Equations
}



\icmlsetsymbol{equal}{*}

\begin{icmlauthorlist}
\icmlauthor{Peng Cui}{thu}
\icmlauthor{Zhijie Deng}{thu}
\icmlauthor{Wenbo Hu}{real}
\icmlauthor{Jun Zhu}{thu}
\end{icmlauthorlist}

\icmlaffiliation{thu}{Dept. of Comp. Sci., Tsinghua University, Beijing, China.}
\icmlaffiliation{real}{RealAI, Beijing, China}

\icmlcorrespondingauthor{Peng Cui}{xpeng.cui@gmail.com}
\icmlcorrespondingauthor{Jun Zhu}{dcszj@tsinghua.edu.cn}

\icmlkeywords{Machine Learning, ICML}

\vskip 0.3in
]



\printAffiliationsAndNotice{} 

\begin{abstract}
It is critical yet challenging for deep learning models to properly characterize uncertainty that is pervasive in real-world environments. 
Although a lot of efforts have been made, such as heteroscedastic neural networks (HNNs), little work has demonstrated satisfactory practicability due to the different levels of compromise on learning efficiency, quality of uncertainty estimates, and predictive performance.
Moreover, existing HNNs typically fail to construct an explicit interaction between the prediction and its associated uncertainty.
This paper aims to remedy these issues 
by developing SDE-HNN, a new heteroscedastic neural network equipped with stochastic differential equations (SDE) to characterize the interaction between the predictive mean and variance of HNNs for accurate and reliable regression. 
Theoretically, we show the existence and uniqueness of the solution to the devised neural SDE.
Moreover, based on the bias-variance trade-off for the optimization in SDE-HNN, we design an enhanced numerical SDE solver to improve the learning stability. 
Finally, to more systematically evaluate the predictive uncertainty, we 
present two new diagnostic uncertainty metrics.
Experiments on the challenging datasets show that our method significantly outperforms the state-of-the-art baselines in terms of both predictive performance and uncertainty quantification, delivering well-calibrated and sharp prediction intervals.
\end{abstract}

\section{Introduction}
\label{introduction}
Deep neural networks (DNNs) are the \emph{defacto} machine learning (ML) tools and significantly outperform human experts in terms of accuracy in various  applications~\cite{deep2015,he2016deep, vaswani2017attention,devlin2018bert}. 
However, considering the ubiquitous uncertainty in the real world, the high accuracy alone is routinely inadequate.
As widely discussed, the accurate and reliable quantification of uncertainty is increasingly important for a variety of ML systems, 
e.g., autonomous driving~\cite{michelmore2020uncertainty},
hydrological forecast~\cite{klotz2020uncertainty},
medical diagnosis~\cite{begoli2019need} and business demand forecast~\cite{zhu2017deep}. 
Typically, the uncertainty of particular concern is classified into two categories: \emph{Aleatoric} uncertainty and \emph{Epistemic} uncertainty a.k.a. data uncertainty and model uncertainty~\cite{Kendall}. 
For regression tasks, the problem of uncertainty quantification mainly reduces to the estimation of prediction intervals (PIs), which is crucial for better-informed decisions~\cite{pearce2018high}. 

Heteroscedastic neural networks (HNNs)~\cite{nix1994estimating} have long been an effective approach for uncertainty estimation in deep regression, which deliver the predictive mean and variance of the observation simultaneously. 
Despite simplicity and scalability, HNNs may exist a bias in fitting the data because of the possible misspecification of model parameters~\cite{dybowski2001confidence}, which lead to mis-calibrated prediction intervals. 
The pioneering works~\cite{Kendall,Lakshminarayanan} to enhance HNNs for more systematical uncertainty modeling typically draw inspiration from Bayesian neural networks (BNNs), while BNNs themselves are not problemless: 1) the high-dimensional parameters and the over-parameterization nature of DNNs pose fundamental challenges for effective posterior inference, thus unsatisfactory performance~\cite{wenzel2020good} and degraded uncertainty estimates~\cite{fort2019deep} are frequently observed; 2) empirical Bayesian methods like Deep Ensemble~\cite{Lakshminarayanan,wenzel2020hyperparameter}, despite more flexible, 
are still restrictive in aspects of computation and memory complexity. Furthermore, unfortunately, rare of the variants of HNNs have succeeded in establishing a principled interaction between the predictive mean and variance, attributed to the conditional independence assumption between the two quantities. 
A direct consequence is that the prediction (i.e., predictive mean) and the associated uncertainty (i.e., predictive variance) cannot interact with each other directly, which contradicts the common sense of decision making.

This work aims to remedy the aforementioned issues of HNNs to realise more accurate and more reliable regression.
Our core insight is that there is an inherent alignment between the predictive mean \& variance in HNNs and the drift \& diffusion coefficients in a stochastic differential equation (SDE).
Based on this insight, we present SDE-HNN, which incorporates the SDE into HNNs to construct an explicit and principled connection between the predictive mean and variance during the process of solving the SDE.  
Thereby, SDE-HNN enables a mutual boosting between the two aspects of predictive distribution during the optimization process, which is not enjoyed by the existing neural SDE methods for uncertainty quantification~\cite{kong2020sde}. 
In theory, we contribute an analysis on the existence and uniqueness of the solution to the devised neural SDE. Moreover, inspired by the bias-variance trade-off, we propose a variant of the Euler-Murayama~\cite{kloden1992numerical} method to discretize and solve the devised SDE, which works in light of dropout~\cite{srivastava2014dropout}.

Intuitively, 
the model uncertainty characterized by the SDE is naturally integrated with the data uncertainty conveyed by the HNNs, yielding an efficient alternative for the combination of BNNs and HNNs. 
For the flexible adoption of the proposed method in practice, we further wrap the building and solving details of the neural SDE to constitute drop-in replacements for DNN blocks.

In practice, we usually expect the prediction intervals (PIs) given by the regressors not only to firmly cover the ground truth (calibration), but also as tight as possible (sharpness). To holistically evaluate the predictive uncertainty of different models, we propose two new diagnostic metrics: the confidence-weighted calibration error (CWCE) and its variant R-CWCE.
Empirically, our method establishes a new state-of-the-art on challenging, large-scale time-series forecasting tasks, presenting remarkable improvement over competitive baselines in terms of both accuracy and uncertainty quantification, and producing high-quality PIs. 






\section{Related Work}
As widely criticized, deep neural networks (DNNs) frequently suffer from over-fitting and are prone to yield over-confident predictions because they characterize the deterministic relationships between the observations and the targets of interest regardless of the other uncertain factors. 
As a compensation, Bayesian deep learning aims at equipping expressive DNNs with appropriate uncertainty quantification~\cite{balan2015bayesian,wang2016towards,Kendall}, with Bayesian neural networks (BNNs) as popular examples.
Typically, BNNs perform the rigorous Bayesian inference over the high-dimensional network weights~\cite{graves2011practical,welling2011bayesian,blundell2015weight,liu2016stein,sun2017learning,louizos2017multiplicative,zhang2018noisy,khan2018fast,osawa2019practical}.
Yet, it is widely observed that BNNs suffer from the difficulty of prior specification~\cite{sun2018functional}, inevitably added training overhead when employing flexible variational posteriors~\cite{shi2018spectral}, as well as the curse of dimensionality when resorting to sampling-based inference~\cite{neal1995bayesian}, which undermine the promise of Bayes principle.
Empirical BNNs, e.g., Monte Carlo (MC) dropout~\cite{gal2016dropout}, Deep Ensemble~\cite{Lakshminarayanan}, SWAG~\cite{maddox2019simple}, and BatchEnsemble~\cite{wen2020batchensemble} enjoy higher scalability and more outperforming predictive performance, but bear undesirable problems like degenerated uncertainty estimates~\cite{fort2019deep}, expensive training/storage cost, and limited expressiveness, etc.
These issues together raise the requirement of a more efficient and more practically reliable uncertainty quantification paradigm to get the accurate and reliable regression.

Inspired by the motion of particles under environmental disturbances in stochastic differential equations (SDE), we suggest constructing the interaction between the prediction and the inherent uncertainty in HNNs to achieve more reliable regression.
Existing work has made a step towards combining SDE with DNNs for uncertainty quantification~\cite{kong2020sde}. However, on the one hand, \citet{kong2020sde} optimize the drift network and diffusion network in a separate, alternative manner, which may be deviated from the Euler-Murayama method for solving SDE and cannot construct the explicit connection between the predictive mean and variance for HNNs. On the other hand, they mainly focus on the out-of-distribution (OOD) detection task based on OOD training data and ignore to explicitly evaluate uncertainty in a principled way. 
The proposed approach differs from it in that we directly work on improving the uncertainty quantification and predictive accuracy for the deep regression by simulating the information flow under data and model perturbation in HNNs using SDEs, and develop an efficient numerical solver to deal with the devised neural SDE. 
\section{Preliminaries}
In this section, we briefly review the background of heteroscedastic neural networks and stochastic differential equations.
\subsection{Heteroscedastic Neural Networks}
To arm DNNs with uncertainty estimation, a straightforward approach is to refine the DNNs to output the prediction value and the associated uncertainty concurrently. 
In this spirit, \citet{nix1994estimating} first introduce the heteroscedastic regression network based on heteroscedastic theory~\cite{engle1982autoregressive}, which assumes a diagonal Gaussian predictive distribution and outputs the mean $\hat{y}$ and variance $\sigma^2$ at the same time:
\begin{equation}
\setlength{\abovedisplayskip}{3pt}
\setlength{\belowdisplayskip}{3pt}
    [\hat{y},\sigma^2] = I(x,\theta),
\end{equation}
where $I$ is a heteroscedastic neural network (HNN) parameterized by $\theta$. 
Nevertheless, there is evidence to suggest that HNNs tend to suffer from the misspecification of model parameters, and hence miscalibrated predictive uncertainty~\cite{dybowski2001confidence}. 
Moreover, although HNNs can be substantially boosted by the advances in Bayesian deep learning~\cite{Kendall,Lakshminarayanan}, most of the variants of HNNs ignore to explicitly characterize the influence of the associated uncertainty on the made prediction, leading to the fact that the predictive mean and variance merely have a weak interaction through the loss function.
Motivated by Brownian motion which is used to model the randomness of particles in physics~\cite{morters2010cambridge}, we argue that it is necessary to explicitly construct a principled interaction between predictive mean and variance for uncertainty quantification in regression problems.
To be specific, we establish the interaction by incorporating stochastic differential equations into HNNs base on the perspective of a stochastic dynamical system.

\subsection{Stochastic Differential Equations}
In order to simulate realistic dynamic systems with uncertainty, we can add some appropriate disturbance to deterministic differential equations, which is represented as:  
\begin{equation}
\frac{d x_{t}}{d t}=f\left(x_{t}, t\right)+g\left(x_{t}, t\right) * \text { noise }.
\setlength{\abovedisplayskip}{7pt}
\setlength{\belowdisplayskip}{7pt}
\label{eqn:de_noise}
\end{equation}
When the noise term connects to Brown motion~\cite{einstein1905motion} -- the standard tool for uncertainty modeling, the above equation boils down to the Itô SDE~\cite{ito1951stochastic}:
\begin{equation}
\label{eqn:3}
\mathrm{d} x_{t}=f\left(x_{t}\right) \mathrm{d} t+g\left(x_{t}\right) \mathrm{d} B_{t}.
\setlength{\abovedisplayskip}{7pt}
\setlength{\belowdisplayskip}{7pt}
\end{equation}
Where $B_t$ denotes a standard Brownian motion, and $f(\cdot)$ and $g(\cdot)$ refer to the drift coefficient and diffusion coefficient respectively. The principled interaction between the motion of particles (i.e., $dx_t$) and environmental disturbances (i.e., $dB(t)$) in the SDE inspires us to construct a connection between the predictive mean and variance in HNNs for regression problems.
\section{SDE-HNN: An Improved Framework for Accurate and Reliable Regression}
In this section, we first describe a framework to incorporate stochastic differential equations into the HNNs to achieve a practical and reliable uncertainty quantification in regression tasks. Then, we give a theoretical guarantee to show the existence and uniqueness of the solution to SDE-HNN (Theorem~\ref{theory:solution}). 
Furthermore, from the perspective of bias-variance trade-off (Theorem~\ref{theory:tradeoff}), we present some theoretical understandings on the solving process of the neural SDE, and design a variant of the Euler-Maruyama method to improve the learning stability. 

\subsection{Reliable Regression via SDE-HNNs}

\textbf{Problem Setting.}\quad Considering a typical regression or forecasting task, we assume access to a dataset $\left\{\left(x^{(i)},  y^{(i)}\right)\right\}_{i=1}^N$, with $x^{(i)}\in \mathbb{R}^{d}$ and ${y^{(i)}} \in \mathbb{R}$ denoting the $d$-dimension data and the target, respectively. The learning objective is to fit a $\Theta$-parameterized regressor $H$: $x \rightarrow y$ according to the data. 
We denote by $p(y|x^{(i)})$ the predictive distribution corresponding to the regressor or forecaster for some data point $x^{(i)}$. 
The cumulative distribution function~(CDF) $F_i$ of such a distribution is informative for quantifying uncertainty -- its inverse $F_i^{-1}:[0,1] \rightarrow \hat{y}^{(i)} $ rigorously serves as the quantile function:
\begin{equation}
F_{i}^{-1}(p)=\inf \left\{y: p \leq F_{i}(y)\right\}.
\setlength{\abovedisplayskip}{4pt}
\setlength{\belowdisplayskip}{4pt}
\end{equation}
In this work, in order to get accurate and reliable regression, we leverage the SDE to construct an explicit connection between predictive mean and variance in HNNs. 
Concretely, we deploy a neural SDE on the hidden representation $z$ of the input $x$:
\begin{equation}
\setlength{\abovedisplayskip}{7pt}
\setlength{\belowdisplayskip}{7pt}
\mathrm{d} z_{t}=f\left(z_{t},\Theta_{f}\right) \mathrm{d} t+g\left(z_{t},\Theta_{g}\right) \mathrm{d} B_{t},
\label{eqn:sde_net}
\end{equation}
where $t$ represents time. 
The functions $f(\cdot)$ and $g(\cdot)$ denote the drift and diffusion networks parameterized by $\Theta_f$ and $\Theta_g$ respectively, which are utilized to estimate the predictive mean and variance, so the interaction between the predictive mean and variance is constructed in the time-continuous dynamic system.
Viewing this dynamic discretely, $z_t$ refers to the state at the $t$-th implicit layer and the parameters are shared among layers. 
\begin{figure*}
    \centering
    \includegraphics[width=0.98\linewidth]{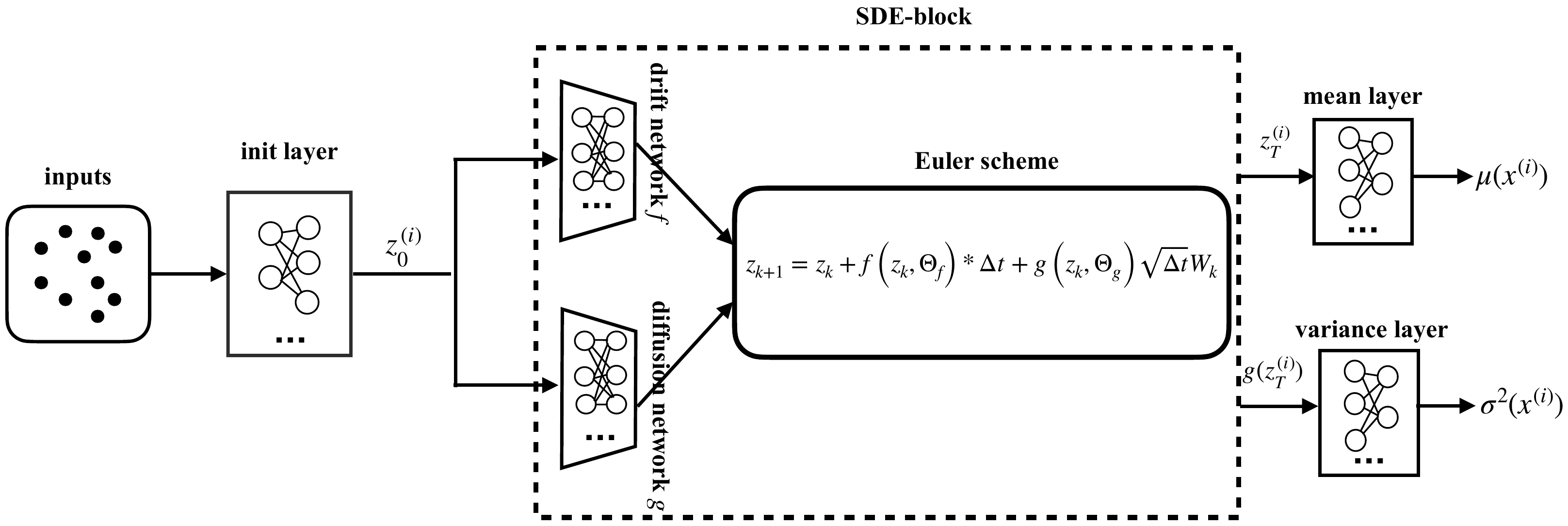}
    \vspace*{-0.4cm}
    \caption{An illustration of SDE-HNN, which mainly includes three parts --- the init layer, the SDE-block and the output layers (i.e., the mean layer and variance layer). The original HNNs do not have the intermediate SDE-block module. The SDE-block can simulate the explicit interaction between the predictive mean and variance during the solving process of SDE.
    }
    \label{fig:framework}
\vspace{-0.3cm}
\end{figure*}

HNNs output the different variance $\hat{\sigma}^2(x^{(i)})$ for each input $x^{(i)}$ base on the heteroscedastic theory~\cite{engle1982autoregressive}. Correspondingly, in our setting, the predictive mean $\mu(x^{(i)})$ and variance $\sigma(x^{(i)})$ can be estimated as follows:
\begin{equation}
\setlength{\abovedisplayskip}{8pt}
\setlength{\belowdisplayskip}{8pt}
\begin{aligned}
&\mu(x^{(i)})=h_1(z^{(i)}_{T},\Theta_{h_1}),\\ 
&\sigma^2(x^{(i)})=h_2(g(z^{(i)}_{T}),\Theta_{h_2}),
\end{aligned}
\end{equation}
where $h_1$ and $h_2$ are two different linear layers (i.e., the mean layer and variance layer in Fig.\ref{fig:framework}) parameterized by $\Theta_{h_1}$ and $\Theta_{h_2}$  respectively and $z^{(i)}_{T}$ is the equilibrium state of the hidden representation for input $x^{(i)}$. 
We drive the neural SDE to fit the data by minimizing negative log-likelihood loss~(NLL):
\begin{equation}
\small
\setlength{\abovedisplayskip}{7pt}
\setlength{\belowdisplayskip}{7pt}
    {L}(\Theta_f;\Theta_g) =\sum_{i=1}^{N}\frac{\log \sigma^{2}({x^{(i)}})}{2}+\frac{\left(y^{(i)}-\mu({x^{(i)}})\right)^{2}}{2 \sigma^{2}({x^{(i)}})}+\text { constant }.
\label{eqn:loss_raw}
\end{equation}
Empirically, to improve numerical stability, we can minimize:
\begin{small}
\begin{equation}
\setlength{\abovedisplayskip}{4pt}
\setlength{\belowdisplayskip}{4pt}
\begin{aligned}
    &{L}(\Theta_f;\Theta_g)=\sum_{i=1}^{N} \frac{\left(y^{(i)}-\mu({x^{(i)}})\right)^{2}}{2} \exp \left(-s^{(i)}\right)+\frac{1}{2} s^{(i)}, 
\label{eqn:loss_log}
\end{aligned}
\end{equation}
\end{small}
where $s^{(i)}:=\log \sigma^{2}(x^{(i)})$. 

\textbf{The Existence and Uniqueness of Solutions}. In order to ensure that the constructed interaction is beneficial for both accuracy and uncertainty quantification in our framework, we need to make sure that the solution to the incorporated SDE exists. Analogous to the previous works on neural SDE~\cite{kong2020sde}, we theoretically show the existence and uniqueness of the solution to the neural SDE in our framework. 
\begin{theorem}
\label{theory:solution}
(Proof in Appendix~\ref{sec:appendix-theory1}) Suppose that $f\left(x_{t},\Theta_{f}\right)$ and $g\left(x_{t},\Theta_{g}\right)$ uniformly satisfy the Lipschitz condition, i.e., there exists a constant $K>0$ making the following inequality hold: 
\begin{equation}
\begin{aligned}
&||f\left({m},{\Theta}_{f}\right)-f\left({n}, {\Theta}_{f}\right)||+||g\left({m},{\Theta}_{g}\right)-g\left({n}, {\Theta}_{g}\right)||\\ &\leq K||m-n||, \quad \forall m, n \in \mathbb{R}^d, t \geq 0.
\end{aligned}
\end{equation}
Then the stochastic differential equations have the unique solution. 
\end{theorem}

Consequently, to make the neural SDE valid and solvable, we need to guarantee that the neural functions $f\left(x_{t}, \Theta_{f}\right)$ and $g\left(x_{t}, \Theta_{g}\right)$ uniformly satisfy the Lipschitz condition.
To this end, we apply the well-evaluated spectral normalization trick~\cite{miyato2018spectral} to the weight matrix in the networks $f(\cdot)$ and $g(\cdot)$. 

\textbf{Solve the Neural SDE}.\quad In general, there is no close-form solution for SDE. 
We need to discretize the SDE and resort to numerical methods to iteratively approach the optimal solution.
The process of solving SDE can actually be viewed as simulating the dynamic corresponding to this SDE. 
A typical approach we can take is the Euler-Murayama method~\cite{kloden1992numerical}, which iteratively applies the following transformation:

\begin{small}
\label{eqn:sde_euler}
${z}_{k+1}={z}_k+f\left({z}_{k}, {\Theta}_{f}\right) (t_{k+1}-t_k)+g\left({z}_{k}, {\Theta}_{g}\right) (B_{t+1}-B_{t})$,
\end{small}

where $k$ is the index of Euler iterations, with $z_0$ as some initial representation of the input $x$. 
According to the properties of the Brownian motion~\cite{durrett2019probability}: $B_{t+1}-B_{t}\sim \mathcal{N}(0, t_{k+1}-t_k)$, then we have $B_{k+1}-B_{k}= \sqrt{\Delta t} W_k$, where $W_k$ follows the standard normal distribution. Therefore, we can rewrite the above transformation:
\begin{equation}
\setlength{\abovedisplayskip}{7pt}
\setlength{\belowdisplayskip}{7pt}
{z}_{k+1}={z}_k+f\left({z}_{k}, {\Theta}_{f}\right) \Delta t+g\left({z}_{k}, {\Theta}_{g}\right) \sqrt{\Delta t} W_{k},
\label{eqn:Euler}
\end{equation}
where $\Delta t$ is the step size of iterations. Let the terminal time of the stochastic process equal to $T$ and the number of iterations equal to $M$, then $\Delta t = T/M $ and the final solution is $z_{T}$. Intuitively, $T$ can be regarded as the number of implicit layers in SDE-HNN.

Our framework is shown in Fig.\ref{fig:framework}, where the Euler solution process can be directly plugged into HNNs as an SDE-block without complicated model modification. The solving process follows the forward and backward propagation of DNNs and can be easily implemented with deep learning libraries. 
The $f(\cdot)$ and $g(\cdot)$ used to estimate the predictive mean and variance are jointly optimized in our framework. 
As a consequence, the prediction and the associated uncertainty can be interacted with each other in hidden features in a principle way. 
In contrast, the separate optimization for $f(\cdot)$ and $g(\cdot)$ in \citet{kong2020sde} cannot construct such an interaction. 

\textbf{Uncertainty Quantification.}\quad Once the model has been trained, we can effectively capture both \emph{Aleatoric} uncertainty and \emph{Epistemic} uncertainty. To be specific, we can get a set of $M$ sampled predictions: $\left\{\boldsymbol{\mu}(x^{(i)}),\boldsymbol{\sigma}^2(x^{(i)})\right\}_{m=1}^{M}$,
and then compute the two kind of uncertainties with the samples via 
\begin{equation}
\nonumber
\setlength{\abovedisplayskip}{4pt}
\setlength{\belowdisplayskip}{0pt}
\label{data_uncertainty}
\emph{Ale}(x^{(i)}) = \frac{1}{M} \sum_{m=1}^{M} \sigma_{m}^{2}(x^{(i)}).
\end{equation}
\begin{equation}
\nonumber
\setlength{\abovedisplayskip}{0.1pt}
\setlength{\belowdisplayskip}{0pt}
\label{model_uncertainty}
\emph{Epi}(x^{(i)}) = \frac{1}{M} \sum_{m=1}^{M} {\mu}_{m}(x^{(i)})^{2}-\left(\frac{1}{M} \sum_{m=1}^{M} {\mu}_{m}(x^{(i)})\right)^{2}.
\end{equation}
\begin{figure}[htbp]
\vspace*{-0.4cm}
\centering
\subfigure[The NLL comparison of \newline two solving methods.]{
\begin{minipage}[t]{0.48\linewidth}
\centering
\includegraphics[width=1\linewidth]{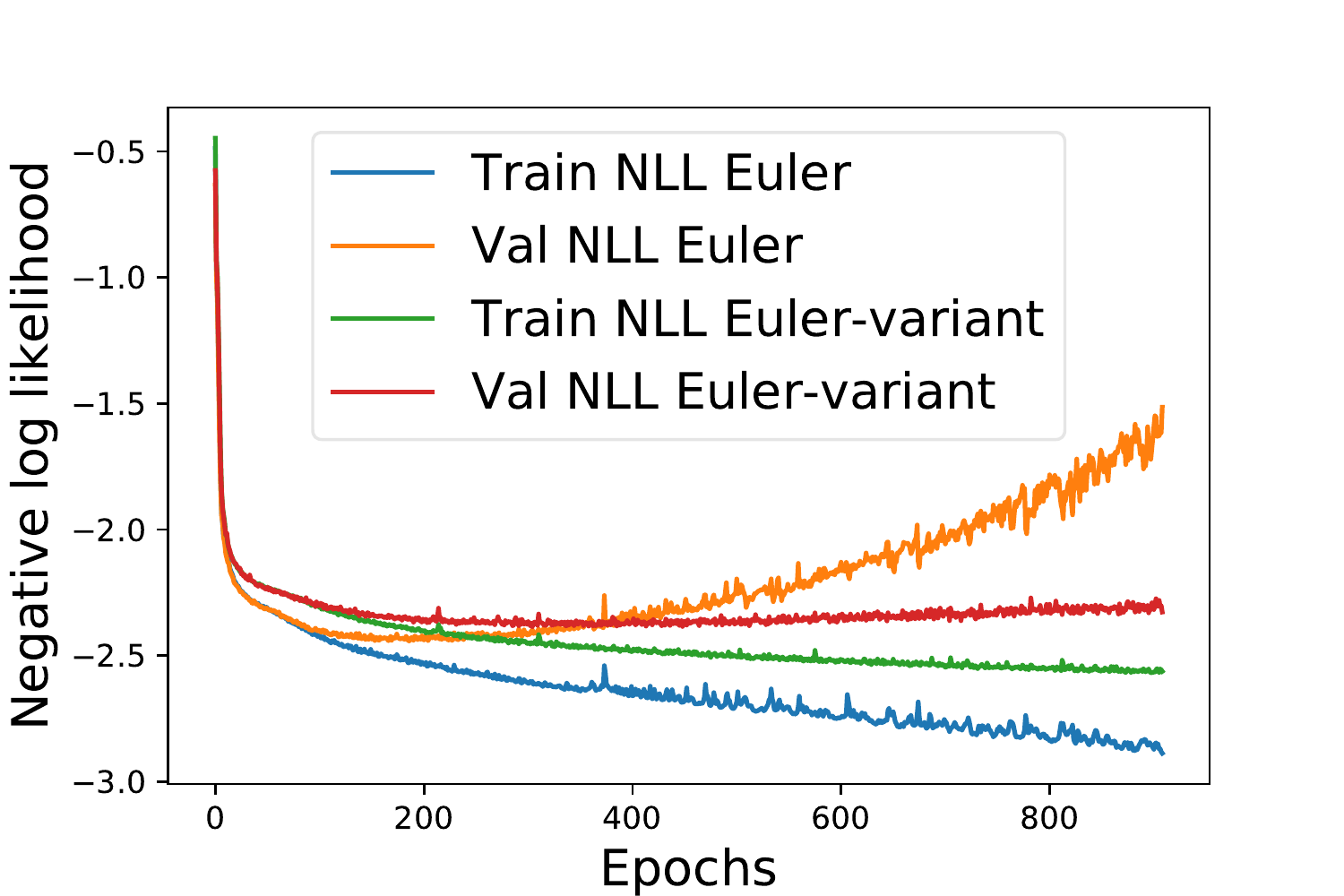}
\end{minipage}%
}%
\subfigure[The CWCE comparison of \newline two solving methods.]{
\begin{minipage}[t]{0.48\linewidth}
\centering
\includegraphics[width=1\linewidth]{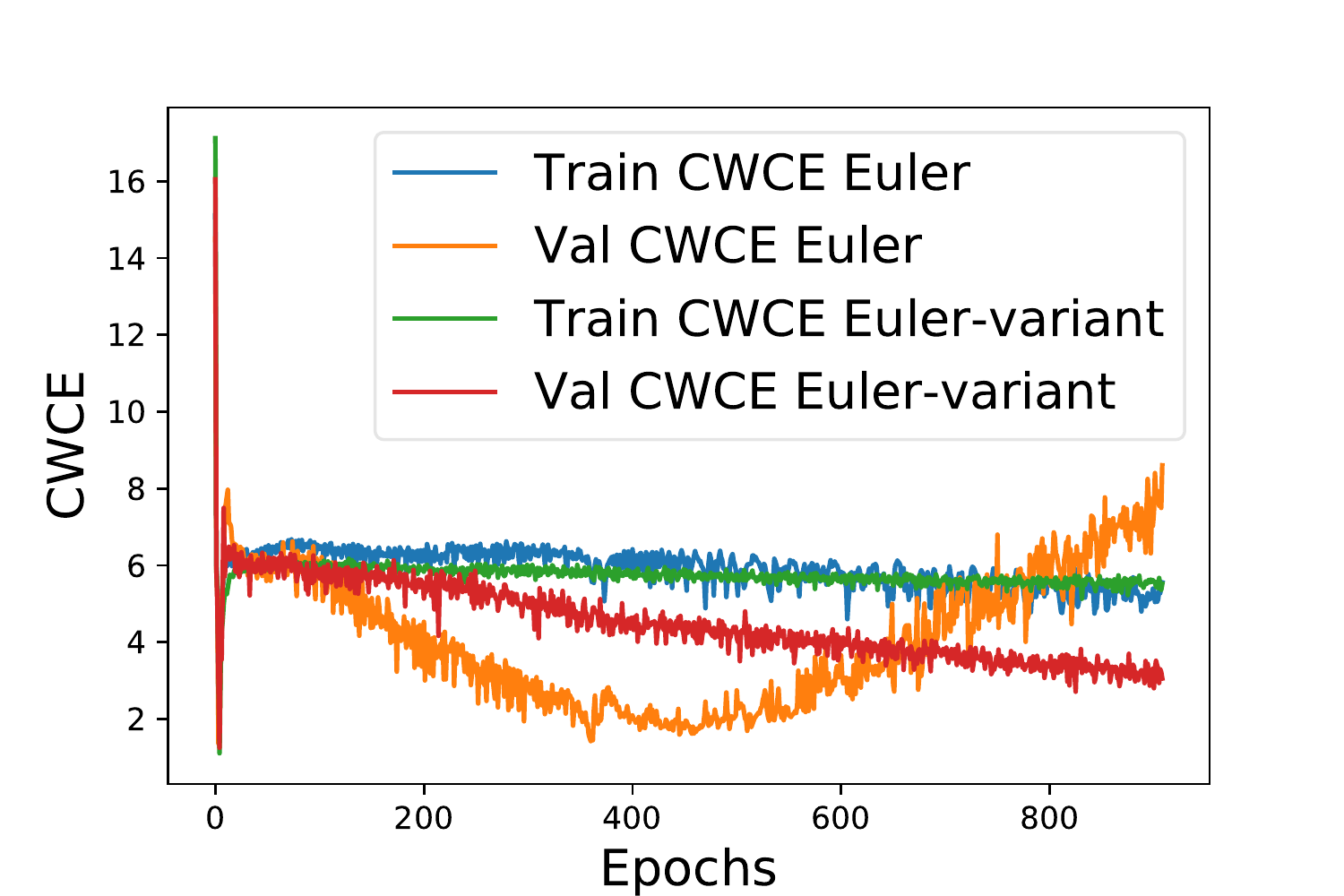}
\end{minipage}%
}
\centering
\vspace*{-0.4cm}
\caption{Plotting the changing curve of NLL and CWCE on training data and validation data on the metro-traffic dataset. The CWCE is a metric of uncertainty evaluation defined in Sec.\ref{sec:cwce}}
\label{fig:Euler-variant}
\vspace*{-0.4cm}
\end{figure}
\subsection{An Improved Solver Motivated By Bias-variance Tradeoff}
We empirically find that the training process may become unstable when the network is close to convergence. 
This phenomenon can be interpreted by the bias-variance tradeoff that exists in the joint optimization process of $f(\cdot)$ and $g(\cdot)$. We give the following analysis and present an improved solver, 
\begin{theorem}
\label{theory:tradeoff}
(Proof in Appendix \ref{sec:appendix-bias-var}) Given a dataset $\left\{\left(x_{i},  y_{i}\right)\right\}_{i=1}^K$ of size $K$, we assume that the training error of the model on this dataset is a constant. In our setting, the quantile function of the model that we want to learn from the dataset is $F^{-1}(p;x)$, dubbed as $\mathcal{F}$ for brevity. There is a bias–variance tradeoff in the process of solving the neural SDE, that is, $\text{MSE}(\mathcal{F},y)=\text{Var}_{(x,y)}(\mathcal{F})+\text{Bias}_{(x,y)}(\E_{(x,y)}(\mathcal{F}),y)^2$
\end{theorem}
Because $\mathcal{F}$ is derived from the distribution function (CDF), and the quantile p is stochastic, the output of $\mathcal{F}$ corresponds to a stochastic output sampled from the SDE at each time $t$.
The Bias$(E(\mathcal{F}),y)^2$ will gradually decrease as the network converges, and then Var$(\mathcal{F})$ will increase because of the constant assumption of MSE$(\mathcal{F},y)$. The variance is mainly controlled by $g(\cdot)$ in our optimization  procedure, so the training process will become unstable because of the increasing $g(\cdot)$. In order to improve the stability of network and make the network asymptotically converge to the optimal point for both mean and variance in the regression, we propose a variant of the standard Euler-Maruyama method.
Specifically, we convert the original deterministic Gaussian uncertainty into Bernoulli's Gaussian uncertainty in the Euler-Maruyama equation to restrain possible explosive $g(\cdot)$, which has the following form:
\begin{equation}
\nonumber
{z}_{k+1}={z}_{k}+f\left({z}_{k}, {\Theta}_{f}\right) \Delta t + \mathcal{D}(g\left({z}_{k}, {\Theta}_{g}\right)) \sqrt{\Delta t} W_{k}.
\label{eqn:Euler_variant}
\end{equation}
Because the diffusion network $g(\cdot)$ has only one layer in our framework, we can simply utilize the Dropout~\cite{srivastava2014dropout} technique to realize the proposed approach in neural networks. Specifically, $\mathcal{D}(\cdot)$ can be regarded as applying dropout to the network $g(\cdot)$, so the Gaussian uncertainty is added with probability $p$ during each Euler iteration, where $p$ is the probability of the neuron being masked. 
Fig.\ref{fig:Euler-variant} shows the changing curve of NLL and CWCE (the CWCE is a metric of uncertainty evaluation defined in Sec.\ref{sec:cwce}) at training time, we can observe that the proposed variant of the Euler-Maruyama method is stabler than the standard Euler-Maruyama and can stably achieve better accuracy and uncertainty estimation performance on the validation dataset. In contrast, the original Euler method may cause the NLL to overfit to the training dataset, resulting in unstable training and poor predictive uncertainty performance. 
\begin{figure}[ht]
\vspace*{-0.3cm}
    \centering
    \includegraphics[width=0.95\linewidth]{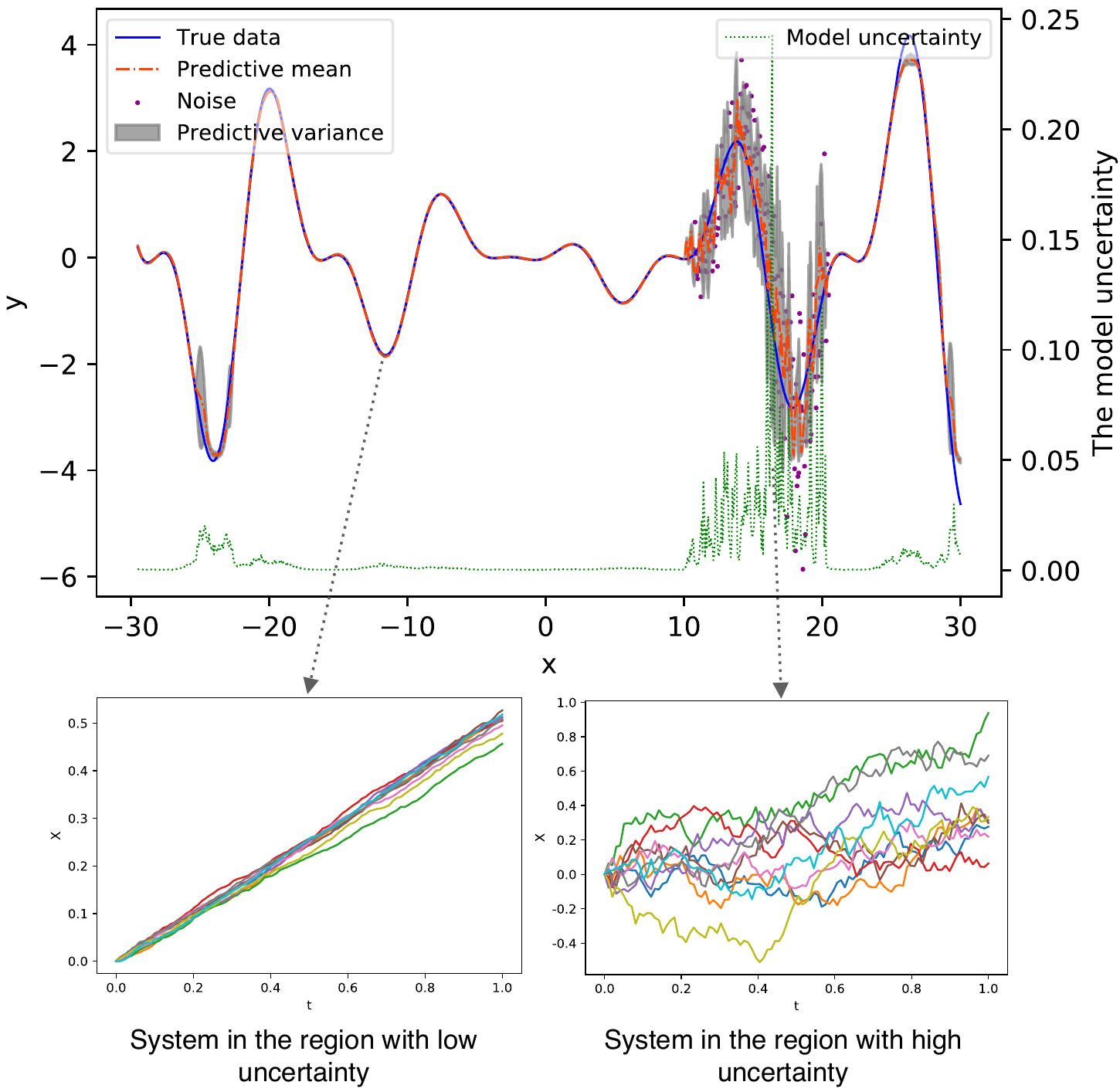}
    \vspace*{-0.4cm}
    \caption{Results on a synthetic dataset, the proposed method can quantify two types of uncertainties simultaneously (the predictive variance corresponds to the data uncertainty). The trajectories of the system are scattered in the region with noise (the right figure at the bottom), oppositely, trajectories of the system are stable in the region with deterministic function (the left figure at the bottom).}
    \label{fig:toy_demo}
\vspace*{-0.4cm}
\end{figure}
\section{Experimental Results}
In this section, we first evaluate the behavior of the proposed method qualitatively on synthetic data. Secondly, we compare the proposed method with heteroscedastic neural networks and other strong baselines on challenging and large-scale time series forecasting tasks in terms of prediction accuracy and predictive uncertainty. It is commonly recognized that time series forecasting is a crucial and complicated task in data science and machine learning, which needs to model multiple regression sub-problems in sequence. Last but not least, we show that the computation efficiency of the proposed method compared with the baseline methods.
\subsection{Toy example}
We firstly qualitatively and visually show the behavior of the proposed method on a one-dimensional synthetic dataset. The synthetic dataset consists of 1000 data points, where the independent variable $x$ is uniformly sampled within $[-30,40]$ and the dependent variable $y$ is obtained via the function $y = (0.4x) \times \sin{x} + (0.7x)\times \cos{x/2}$. Moreover, we add the heteroscedastic Gaussian noise $\epsilon_t$ within $[10,20]$ via the function to observe the behavior of the dynamic system and network, where $\epsilon \sim \mathcal{N}(0,0.0225x^2)$. Knowing the true noise in our synthetic dataset, we can evaluate the quality of predictive uncertainty. 

We use the model with 64 units in all hidden layers (the init layer, drift network $f(\cdot)$ and diffusion network $g(\cdot)$). The terminal time $T$ and step size $\Delta t$ of Euler solver are set to 3 and 1 respectively. 
The results are shown in Fig.\ref{fig:toy_demo}. We can observe that the predictive variance (data uncertainty) becomes higher in the area where heteroscedastic Gaussian noise is added within $[10,20]$ and the model uncertainty can increase within $[10,20]$ because the made prediction is not too exact in the area. The two figures in Fig.\ref{fig:toy_demo} at the bottom show that ten trajectories sampled from another linear SDE. If the system is in the region with clean data and deterministic pattern, the system is stable. On the contrary, if the system is in the region with noise, the diffusion coefficient is high, the system becomes disordered. 
\subsection{Setup}
\textbf{Datasets.}\quad We conduct our experiments on public multivariate time series data from UCI repository~\cite{Dua:2019}, Kaggle~\cite{Yannis:2017} and \citet{lai2018modeling}. A multivariate time series data consists of multiple time-dependent variables, it is more challenging due to the complex mixtures of temporal patterns and inter-dependencies among multivariate time series. Table \ref{tab:label_dataset} describes the details of the datasets in Appendix~\ref{sec:appendix-dataset}.

\textbf{Baselines.}\quad We compare the proposed method with several competitive and popular baselines, including heteroscedastic neural network~(HNN)~\citep{Kendall}, MC-Dropout~(MCD)~\citep{gal2016dropout},  Deep Ensembles~(Deep-ens)~\citep{Lakshminarayanan}, Deep Gaussian Processes~(DGP)~\cite{salimbeni2017doubly} and BNN~\cite{blundell2015weight}.

\textbf{Model setting and hyperparameters.}\quad In our experiments, we split each of the six datasets into training data (60\%), validation data (20\%) and testing data (20\%). 
For the model architecture, we construct an LSTM model with two hidden layers (both 128 hidden units) and two linear layers to make the final predictions for all baselines. Fairly, the init layer of the proposed method is one-layer LSTM with 64 hidden units, the drift network $f$ and the diffusion network $g$ are both one-layer LSTM with 64 hidden units. The terminal time $T$ and step size $\Delta t$ of the Euler solver are set to 3 and 0.5 respectively. 

For the data preprocessing, we scale the raw data into the range [0,1] to improve the computation stability and efficiency. And we choose the size of the sliding window is 5 and the forecasting horizon is 1 according to validation datasets. Take the Metro-traffic dataset as an example, the traffic volume data of the past five hours is used to predict the data of one hour in the future.
For optimization algorithm, we all use the Adam~\citep{kingma2014adam} with learning rate $10^{-3}$ and weight decay $10^{-3}$. 
For all experimental results, we report the averaged results obtained from 5 random trials.

\subsection{Evaluation metrics.}
\label{sec:cwce}
Formally, we say a regressor is \emph{well-calibrated}~\cite{dawid1982well,Kuleshov} if for $\forall p\in[0,1]$, the following condition always holds:
\begin{equation}
\frac{\sum_{i=1}^{N} \mathbb{I}\left\{y_{i} \leq F_{i}^{-1}(p)\right\}}{N} \rightarrow p, \text{when } N \rightarrow\infty,
\label{eqn:cali_single}
\end{equation}
where $\mathbb{I}(\cdot)$ is the indicator function that equals to $1$ if the predicate holds
otherwise $0$. The left terms in Eqn.~(\ref{eqn:cali_single}) is generally referred to as the empirical coverage probability, dubbed as $E(p)$. 

In the previous literature, for real-world applications of machine learning, the predictive uncertainty is evaluated from two aspects: calibration and sharpness~\cite{gneiting2007probabilistic}. 
\begin{table*}[!ht]

\vskip -0.2cm
    \centering
    \setlength{\tabcolsep}{2.5mm}{
    \begin{tabular}{|l|l|r|r|r|r|r|r|}
\hline
Dataset & Metric &MCD &DGP  & BNN& Deep-ens &HNN & Proposed\\ \hline
\multirow{3}{*}{Metro-traffic}
 & RMSE   &697.021  &651.341  &786.694  &533.426   &559.354 &
 \textbf{483.639$\pm$2.657}\\  
 & $R^2 \uparrow$   &0.877  &0.892  &0.843  &0.928  &0.920 & \textbf{0.939$\pm$0.011}\\
 & CWCE   &52.152   &10.552  &21.486  &9.078  &9.305 &\textbf{2.894$\pm$0.085}\\
 & EPIW   &\textit{\sout{167.859}}  &1168.044 &610.662  &814.143  &883.475 & \textbf{539.254$\pm$19.334}\\
 & R-CWCE   &6.428  &1.136 &3.373  &0.655  &0.747 & \textbf{0.177$\pm$0.014}\\
 \hline
\multirow{3}{*}{Pickups} 
 & RMSE   &625.812 &523.041 &720.013  &428.032   &421.752 & \textbf{340.331$\pm$5.072}\\  
 & $R^2 \uparrow$   &0.878 &0.914 &0.838  &0.943 &0.945 & \textbf{0.964$\pm$0.012}\\
 & CWCE   &34.441  &22.799 &42.570  &4.878 &6.043 & \textbf{2.925$\pm$0.758}\\
 & EPIW   &\textit{\sout{313.432}} &1872.481 &\textit{247.229} &684.381  &688.989 & \textbf{438.324$\pm$19.222}\\ 
 & R-CWCE   &4.205 &1.951 &6.904 &0.280  &0.335 & \textbf{0.173$\pm$0.012}\\
 \hline
\multirow{3}{*}{Stock} 
 & RMSE   &4.947 &3.530 &8.663  &2.122   &\textbf{1.903} & 2.102$\pm$0.024\\  
 & $R^2 \uparrow$   &0.993  &0.997 &0.981  &0.999  &0.998 & \textbf{0.999$\pm$0.000}\\
 & CWCE   &5.869  &26.911 &15.233  &14.678  &18.430 & \textbf{3.717$\pm$0.069}\\
 & EPIW   &6.421  &15.839  &29.882  &4.273  &4.208 & \textbf{3.303$\pm$0.058}\\ 
 & R-CWCE   &0.039 &0.084 &0.285 &0.017  &0.020 & \textbf{0.004$\pm$0.000}\\
 \hline
 \multirow{3}{*}{Electricity}
 & RMSE   &3780.973 &2936.760    &6106.273  &1535.261   &1592.815 & \textbf{1497.891$\pm$10.253}\\  
 & $R^2 \uparrow$   &0.642  &0.739   &0.415  &0.887  &0.884& \textbf{0.894$\pm$0.003}\\
 & CWCE   &49.138   &9.225   &47.096  &5.575  &4.169 & \textbf{3.555$\pm$0.297}\\
 & EPIW   &\textit{\sout{156.028}}   &492.443   &\textit{\sout{345.186}}  &382.168  &379.107 & \textbf{358.012$\pm$4.778}\\ 
 & R-CWCE   &17.585   &2.405   &27.571  &0.630  &0.486 & \textbf{0.378$\pm$0.005}\\
 \hline
\multirow{3}{*}{Solar}
 & RMSE   &3.405 &2.995  &3.813  &1.982   &2.201 & \textbf{1.940$\pm$0.006}\\  
 & $R^2 \uparrow$   &0.890 &0.938   &0.855  &0.965  &0.964 & \textbf{0.966$\pm$0.001}\\
 & CWCE   &62.528  &9.103   &9.1840  &18.568  &12.081 & \textbf{8.403$\pm$1.684}\\
 & EPIW   &\textit{\sout{0.296}}  &2.996  &2.650 &1.749  &1.946 & 2.418$\pm$0.222\\ 
 & R-CWCE &6.890  &0.569   &1.334  &0.657  &0.439 & \textbf{0.288$\pm$0.002}\\
 \hline
 \multirow{3}{*}{Traffic flow}
 & RMSE   &0.040 &0.047  &0.047  &0.023   &0.024 & \textbf{0.022$\pm$0.000}\\  
 & $R^2 \uparrow$   &0.393  &0.213   &0.112  &0.786  &0.782 & \textbf{0.813$\pm$0.000}\\
 & CWCE   &55.490   &27.229   &46.959  &14.554  &12.706 & \textbf{10.464$\pm$0.931}\\
 & EPIW   &\textit{\sout{0.004}}  &0.030   &\textit{\sout{0.011}}  &0.024  &0.024 & \textbf{0.021$\pm$0.001}\\ 
 & R-CWCE   &33.700  &21.419   & 41.688   &3.109 &2.766 &\textbf{1.956$\pm$0.001}\\
 \hline
\end{tabular}}
\vspace*{-0.25cm}
    \caption{The forecast accuracy, calibration error and sharpness of uncertainty evaluation for each method on different datasets. Each column represents the performance of a specific method for different metrics on different datasets. The larger values represent better performance in the metric with '$\uparrow$'. Note that when the corresponding CWCE and RMSE are too large, EPIW has no comparative meaning with other methods and these EPIWs are written in \textit{\sout{italics}}.}
    \label{tab:results_forecast}
\vskip -0.3cm
\end{table*}

\textbf{Proposed confidence-weighted calibration metric.}
\quad The calibration metrics are traditionally defined through the quantile function for regression problems. In the previous works~\cite{dawid1982well,Kuleshov,cui2020calibrated}, the calibration error is the equivalent accumulation of different quantile (confidence) deviation, which may be inaccurate when the data distribution is uneven. For the realistic usage, the quantile deviation with a large confidence is more important than those with a small confidence.
To overcome this issue, we propose a new metric, confidence-weighted calibration error (CWCE), which is defined as:
\begin{equation}
    \setlength{\abovedisplayskip}{4pt}
    \setlength{\belowdisplayskip}{4pt}
    \text{CWCE} = \sum_{k=1}^{n}p_k\times\left|E(p_k)-p_k\right|,
\end{equation}
where $E(q)$ is the empirical coverage probability and $p$ is the true confidence (i.e., the confidence level that we expect).
\begin{figure}[htbp]
\vspace*{-0.4cm}
\setlength{\abovecaptionskip}{0.cm}
\centering
\subfigure{
\begin{minipage}[t]{0.48\linewidth}
\centering
\includegraphics[width=1\linewidth]{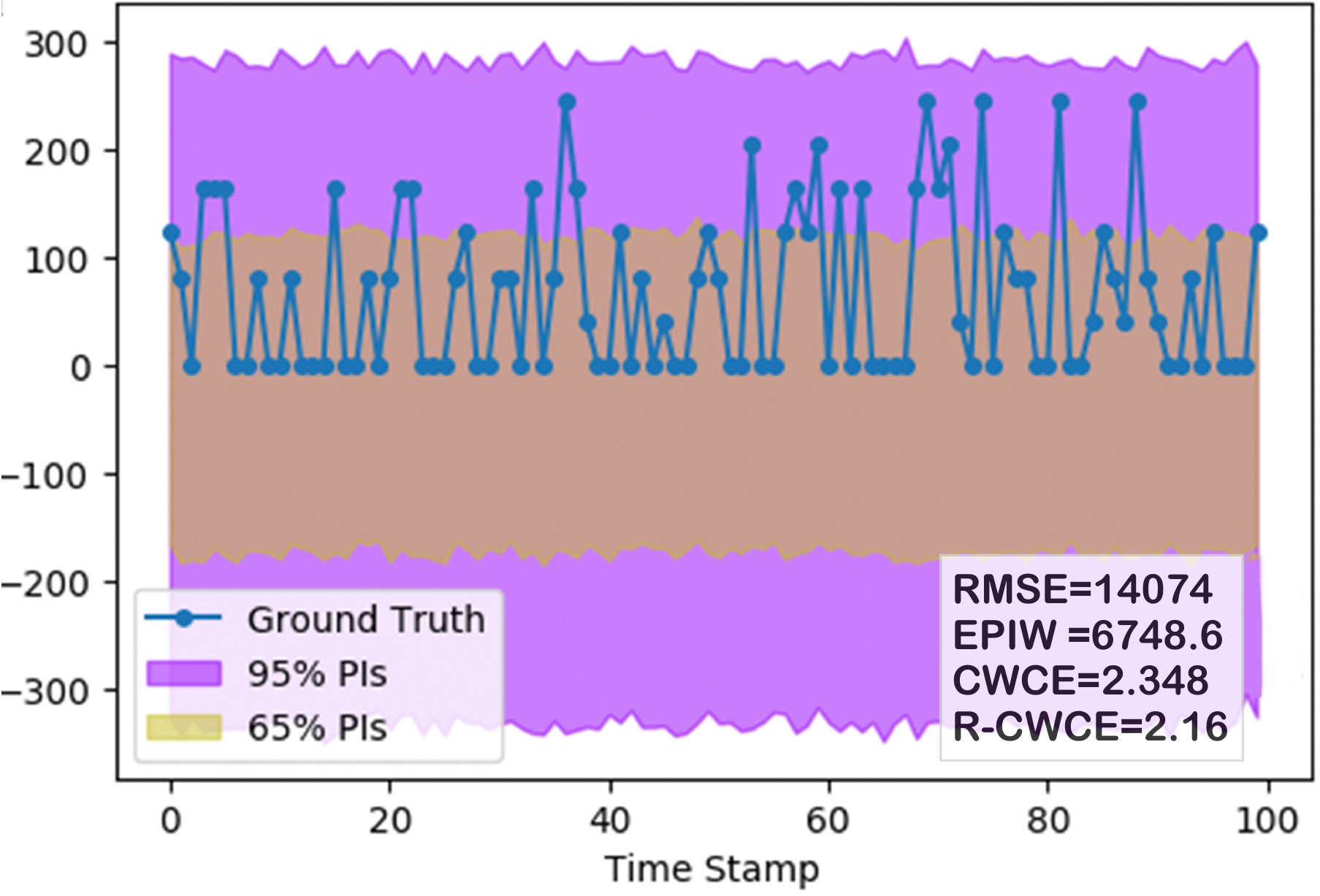}
\end{minipage}%
}%
\subfigure{
\begin{minipage}[t]{0.48\linewidth}
\centering
\includegraphics[width=1\linewidth]{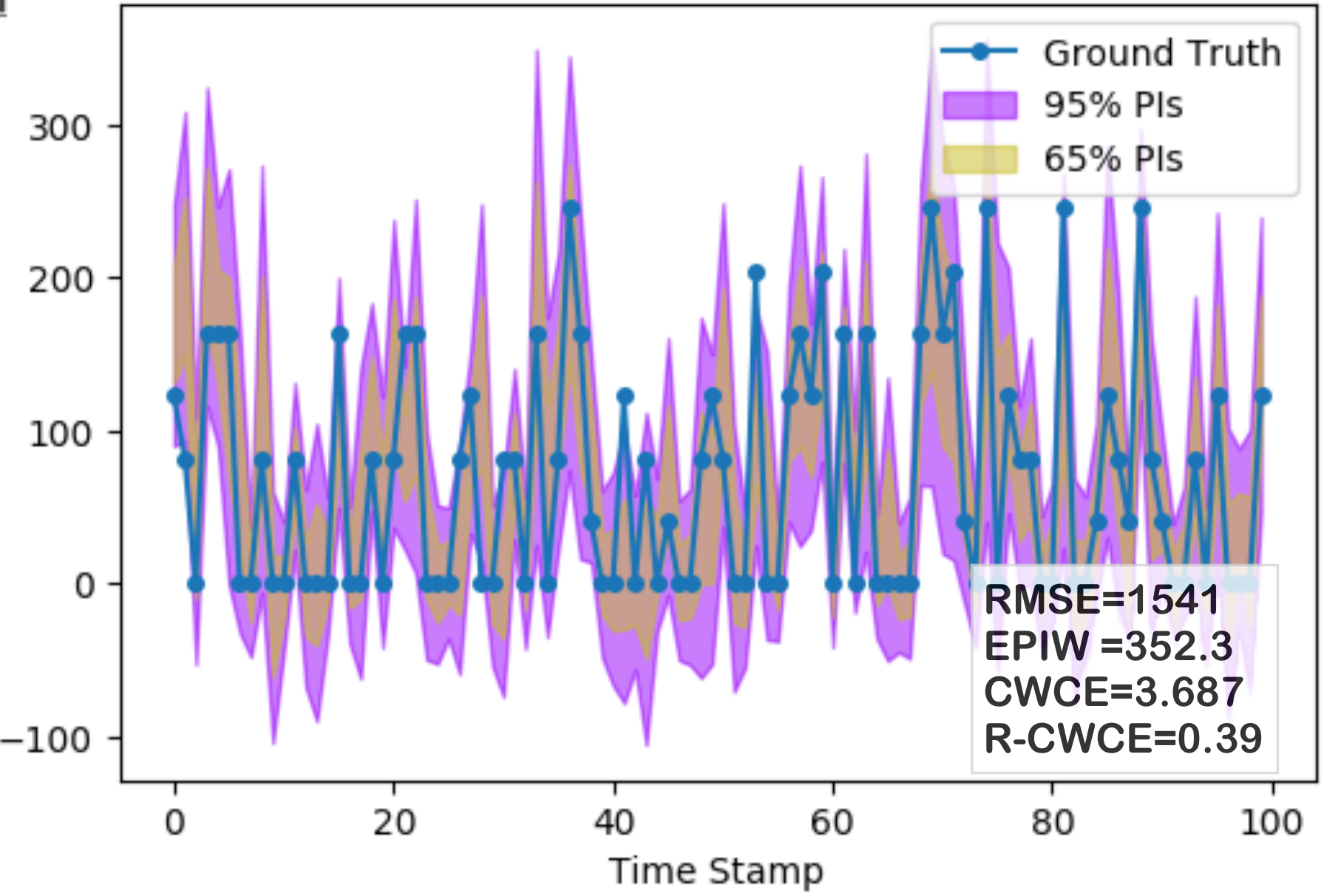}
\end{minipage}%
}
\centering
\vspace*{-0.4cm}
\caption{The partial prediction intervals comparison at the 65\% and 95\% confidence on the electricity dataset. The CWCE and EPIW are equal to $2.348$ and $6748.6$ in the left figure, and the CWCE and EPIW are equal to $3.687$ and $352.3$ in the right figure. \emph{The definitions of EPIW can be found in Appendix \ref{sec:appendix-metrics}.}}
\label{fig:sharpness_compare}
\vspace*{-0.5cm}
\end{figure}

\textbf{Proposed metric for fusing calibration and sharpness.}\quad
Generally, we prefer prediction intervals are as tight as possible when accurately covering the ground truths (i.e., the calibration error is low). At the moment, the sharpness should be considered as a diagnostic tool for uncertainty, which refers to the concentration of the predictive distributions~\cite{gneiting2007probabilistic}. The calibration and sharpness are evaluated separately in the previous works~\cite{Kuleshov,pearce2018high,cui2020calibrated}, so we need to consider both metrics respectively when comparing the results.
For convenience, in this paper, we design a combinative metric to evaluate the calibration and sharpness together. 
Motivated by the coefficient of determination, we propose R-CWCE, the variant of CWCE:
\begin{equation}
    \setlength{\abovedisplayskip}{4pt}
    \setlength{\belowdisplayskip}{4pt}
    \text{R-CWCE} = \frac{\sum_{i}\left(y_{i}-\hat{y_{i}}\right)^{2}}{\sum_{i}\left(y_{i}-\bar{y}\right)^{2}} \times \text{CWCE},
    \end{equation}
where $\hat{y_i}$ is the predictive mean of the model, $\bar{y}$ is the mean of the observed data, the first term of the right of the equation represents the predictive performance of the model.
A smaller R-CWCE represents better calibration and sharpness performance, the value of R-CWCE on the left (\emph{2.16}) is obviously greater than that (\emph{0.39}) on the right in Fig.\ref{fig:sharpness_compare}, which shows that the proposed metric is more effective.

In summary, the performance of all methods is evaluated from two aspects, the precision of prediction and the reliability of uncertainty: 1) the metrics of prediction precision: RMSE and $R^2$, 2) the metrics of the calibration and sharpness: CWCE, R-CWCE and EPIW, where the definitions of RMSE, $R^2$ and EPIW can be found in Appendix \ref{sec:appendix-metrics}.
\begin{figure}[ht]
\vspace*{-0.32cm}
    \centering
    \includegraphics[width=0.9\linewidth]{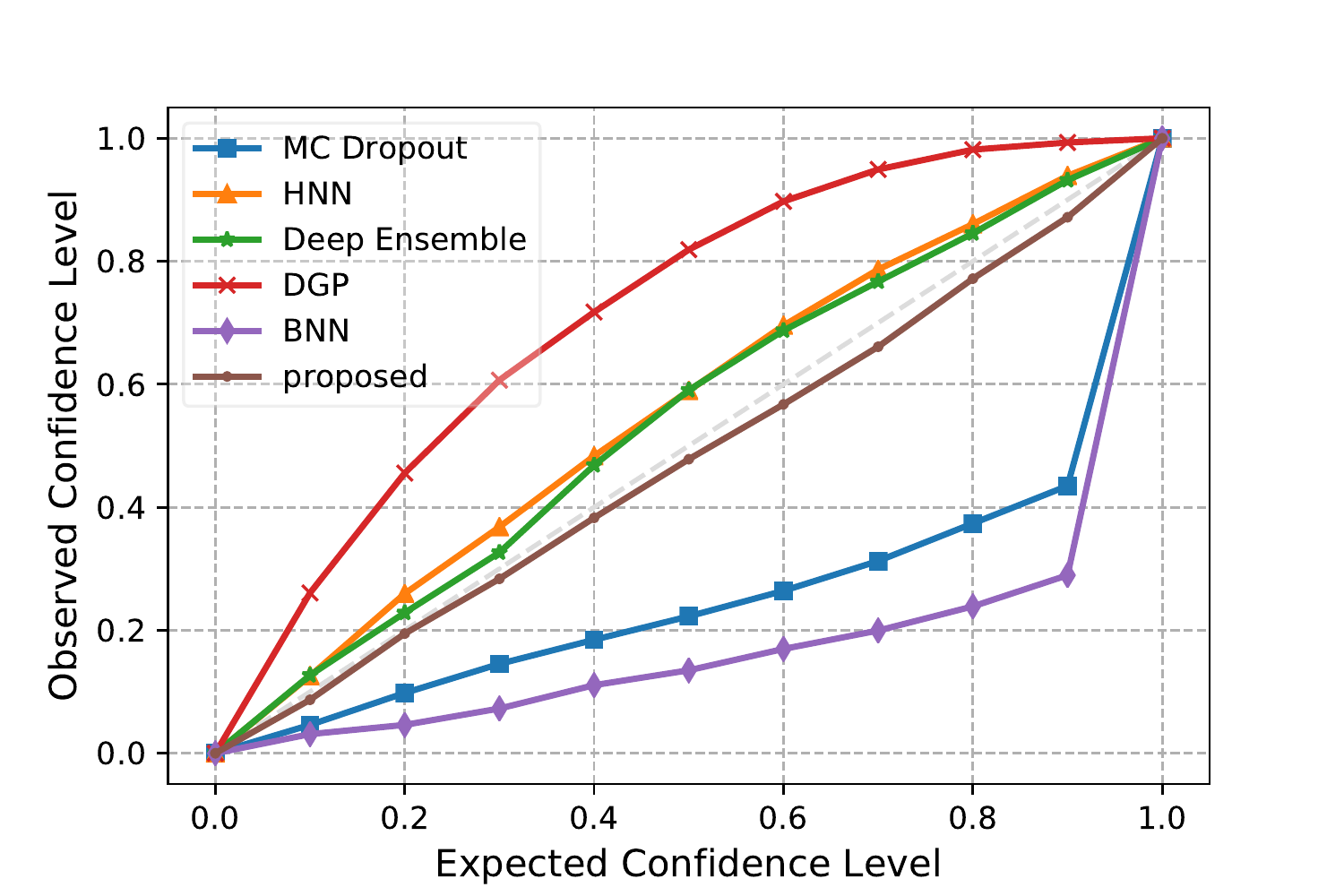}
    \vspace*{-0.5cm}
    \caption{Plots of the empirical coverage probability vs expected confidence for all methods on Pickups dataset.}
    \label{fig:cali_error}
\vspace*{-0.4cm}
\end{figure}
\begin{figure}[ht]
\vspace*{-0.4cm}
    \centering
    \includegraphics[width=0.9\linewidth]{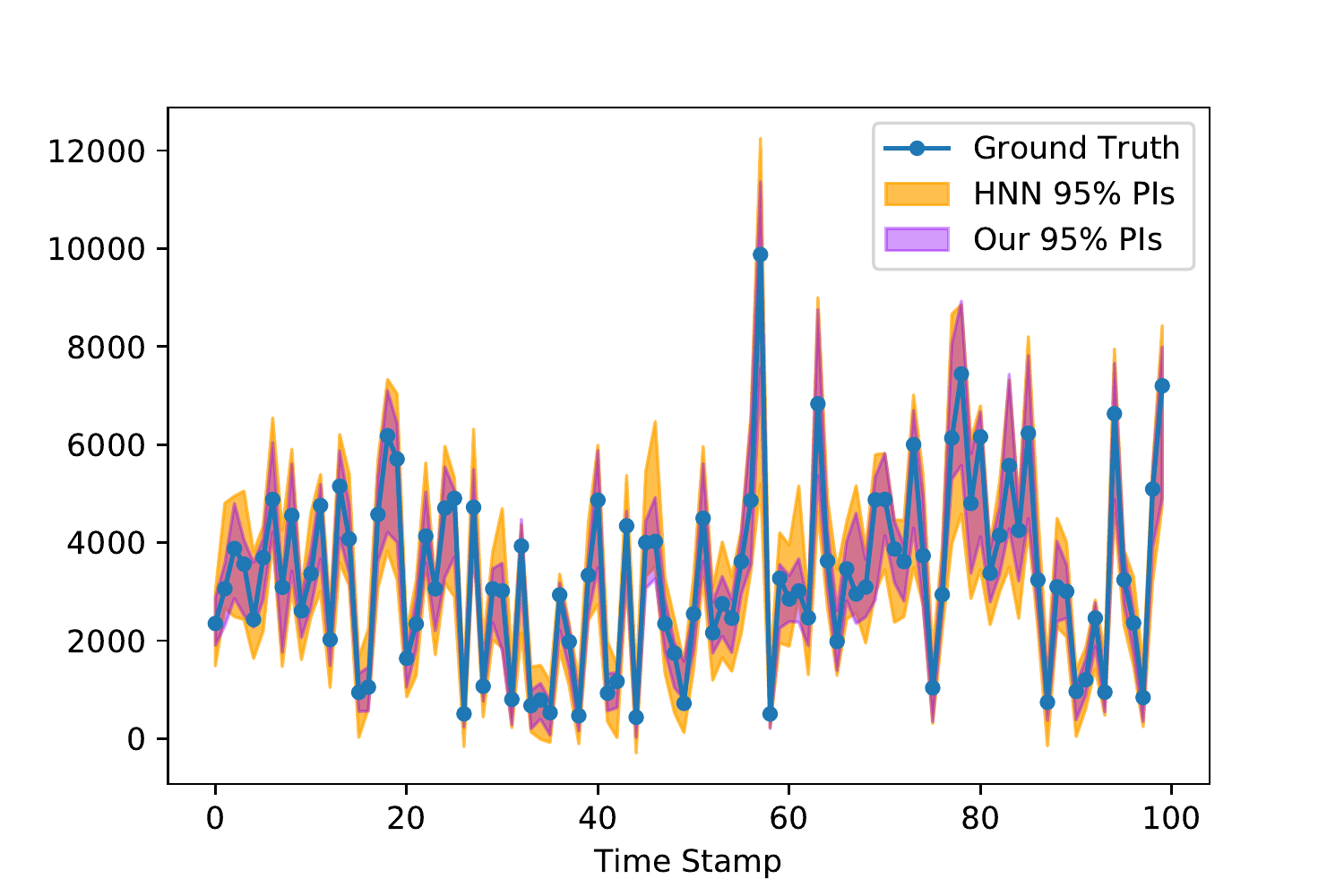}
    \vspace*{-0.5cm}
    \caption{The comparison plots of prediction intervals between HNN and our method on Pickups dataset.}
    \label{fig:pis}
\vspace*{-0.3cm}
\end{figure}
\subsection{Time Series Forecasting on real world datasets}
\label{sec:forecast}
Table~\ref{tab:results_forecast} shows the performance of prediction precision and uncertainty reliability for all methods. We can observe that the proposed method significantly outperforms the baselines in terms of the metrics of forecasting and uncertainty estimation by achieving lower RMSE and calibration error (CWCE). And the R-CWCE of the proposed method is the smallest on all datasets, which shows the effectiveness of the proposed method for both the calibration and sharpness. Especially, the proposed method achieves a great performance improvement in both forecasting and uncertainty estimation compared to HNNs and also greatly outperforms competitive Deep-Ensembles. In order to visually demonstrate the performance of uncertainty estimation, we report calibration curves and prediction intervals in Fig.\ref{fig:cali_error} and Fig.\ref{fig:pis}. Fig.\ref{fig:cali_error} shows the empirical coverage probability at different confidence levels, the result of the proposed model is closest to the expected confidence level, which indicates the best calibration performance among all methods. Fig.\ref{fig:pis} shows 95\% prediction intervals obtained by the proposed method and HNNs, the intervals are visually sharper and can accurately cover the ground truths, which show the proposed method can output well-calibrated and sharp prediction intervals.
\subsection{Computation Efficiency}
\begin{figure}[ht]
\vskip -0.3cm
    \centering
    \includegraphics[width=0.9\linewidth]{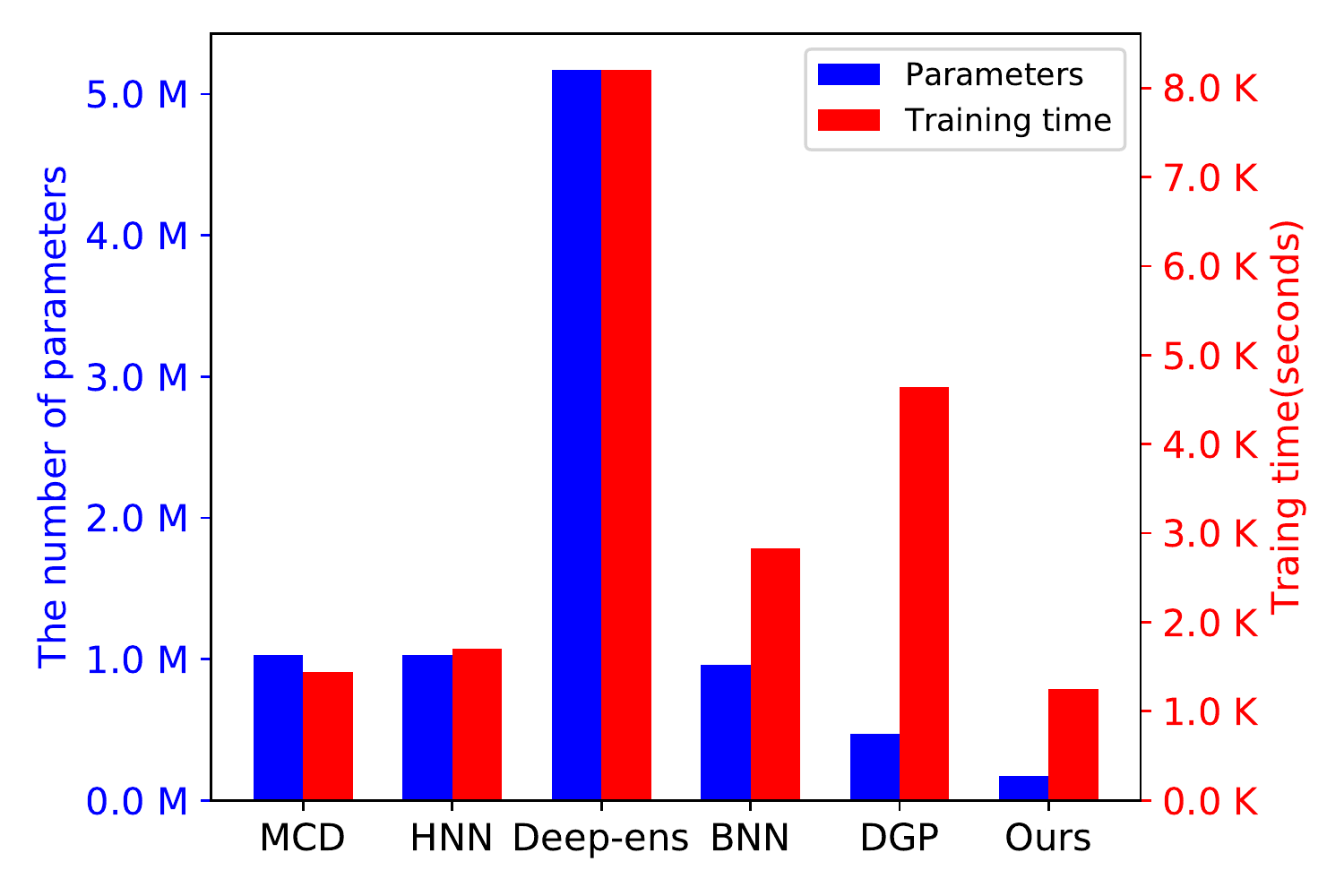}
    \vspace*{-0.5cm}
    \caption{The number of parameters and training time of all methods on stock dataset on GTX1080Ti.}
    \label{fig:efficiency}
\vskip -0.3cm
\end{figure}

We analyze the computational efficiency of all methods on the stock dataset, including the number of parameters and training time. The experimental settings are consistent with Sec.~\ref{sec:forecast}. 
As can be seen in Fig.\ref{fig:efficiency}, Deep Ensemble has the maximum number of parameters and is the most time-consuming. Bayesian methods (i.e., DGP and BNNs) also have a high computational cost. By contrast, the training time and parameters of the proposed method are the least among all methods.
\section{Conclusions and Future Work}
In this work, we present SDE-HNN, a practical and efficient method that incorporates stochastic differential equations (SDE) to explicitly characterize the interaction between the predictive mean and variance of HNNs. 
We show the existence and uniqueness of the solution for our method, provide an analysis for the optimization process based on bias-variance trade-off, and further design a variant of the Euler-Maruyama method to improve the learning stability. Finally, we reexamine the current uncertainty evaluation and propose two new diagnostic metrics. Empirically, our method significantly outperforms the competitive baselines in terms of both predictive accuracy and uncertainty quantification on the non-trivial real datasets.

Because the plug-and-play \emph{SDE-block} is sufficiently flexible in our framework, in future work, we will be devoted to promoting the proposed method and framework to more tasks to get well-calibrated predictions, such as classification and detection. Another worthwhile investigation is the combination of our method and other calibration methods, such as temperature scaling~\cite{guo2017calibration}, non-parametric isotonic regression~\cite{Kuleshov} and distribution matching~\cite{cui2020calibrated}, to further improve the uncertainty quantification in classification and regression tasks.



\bibliography{example_paper}
\bibliographystyle{icml2021}

\clearpage
\appendix
\section{The proof of Theorem 1}
\label{sec:appendix-theory1}
Theorem 1 can be seen as a special case of the existence and uniqueness theorem of a general stochastic differential equation, we first introduce some lemmas.
\begin{lemma}
\label{lemma:borel}
Let $C_1,C_2,\cdot$ be a sequence of events in the probability space. The \emph{Borel–Cantelli} lemma states:

If the sum of the probabilities of the event $E_n$ is finite,
\begin{equation}
    \sum_{n=1}^{\infty} \operatorname{Pr}\left(C_{n}\right)<\infty,
\end{equation}
then the probability that infinitely many of them occur is 0, that is,
\begin{equation}
\operatorname{Pr}\left(\limsup _{n \rightarrow \infty} C_{n}\right)=0.
\end{equation}
\end{lemma}
In the theorem 1, we declare that the stochastic differential equations have the unique solution,
\begin{equation}
\label{appendix:eqn_x}
\mathrm{d} z_{t}=f\left(z_{t} ; t ; \Theta_{f}\right) \mathrm{d} t+g\left(z_{t} ; t ; \Theta_{g}\right) \mathrm{d} B_{t}, z_0 = z.
\end{equation}
Firstly, we prove the uniqueness of solutions. Suppose $y$ is another solution, that is,
\begin{equation}
\label{appendix:eqn_y}
\mathrm{d} y_{t}=f\left(y_{t} ; t ; \Theta_{f}\right) \mathrm{d} t+g\left(y_{t} ; t ; \Theta_{g}\right) \mathrm{d} B_{t}, y_0 = z.
\end{equation}
Then we calculate the square of the subtraction of Eqn.\ref{appendix:eqn_y} and Eqn.\ref{appendix:eqn_x}, then use the basic inequality $(a+b)^{2} \leq 2 a^{2}+2 b^{2}$ and take the expectation, we can get,
\begin{equation}
\begin{aligned}
\mathbb{E}\left|y_{t}-z_{t}\right|^{2} \leq & 2 \mathbb{E}\left(\int_{0}^{t}\left(f\left(y_{s}\right)-f\left(z_{s}\right)\right) d B_{s}\right)^{2} \\
&+2 \mathbb{E}\left(\int_{0}^{t}\left(g\left(y_{s}\right)-g\left(z_{s}\right)\right) d s\right)^{2} \\
\leq & 2 \mathbb{E} \int_{0}^{t}\left|f\left(y_{s}\right)-f\left(z_{s}\right)\right|^{2} d\langle B, B\rangle_{s} \\
&+2 \mathbb{E}\left(t \int_{0}^{t}\left|g\left(y_{s}\right)-g\left(x_{s}\right)\right|^{2} d s\right).
\end{aligned}
\end{equation}
According to the Lipschitz condition, we can get,
\begin{equation}
\small
    \mathbb{E}\left|y_{t}-z_{t}\right|^{2} \leq 2 K^{2} \mathbb{E} \int_{0}^{t}\left|y_{s}-z_{s}\right|^{2} d s+2 K^{2} t \mathbb{E} \int_{0}^{t}\left|y_{s}-z_{s}\right|^{2} ds.
\end{equation}
Let $c(t)=\mathbb{E}\left|y_{t}-z_{t}\right|^{2}$, then for any $t \leq T$, we can get 
\begin{equation}
c(t) \leq C_{T} \int_{0}^{t} c(s) d s,
\end{equation}
where $C_{T}=2 K^{2}(1+T)$, from the above inequality, we can immediately conclude that $c(t)=0$, so the uniqueness holds.

Moreover, we use iterative methods to prove the existence and then define the following equation:
\begin{equation}
z_{t}^{n}=z+\int_{0}^{t} f\left(z_{s}^{n-1}\right) d s+\int_{0}^{t} g\left(z_{s}^{n-1}\right) d s.
\end{equation}
We will prove the convergence of $z^n$. We take the subtraction of $z_t^{n+1}$ and $z_t^n$, and then we can get,
\begin{equation}
\small
\begin{aligned}
\max _{0 \leq s \leq t}\left|z_{s}^{n+1}-z_{s}^{n}\right| \leq & 2 \max _{0 \leq s \leq t}\left(\int_{0}^{s}\left[f\left(z_{u}^{n}\right)-f\left(z_{u}^{n-1}\right)\right] d u\right)^{2} \\
&+2 T K^{2} \int_{0}^{t} \max _{0 \leq u \leq s}\left|z_{u}^{n}-z_{u}^{n-1}\right| d s.
\end{aligned}
\end{equation}
Let $\eta_{n}(t)=\mathbb{E}\left[\max _{0 \leq s \leq t}\left|z_{s}^{n}-z_{s}^{n-1}\right|^{2}\right]$, from the moment inequality of Doob martingale, for any $t\leq T$, 
\begin{equation}
\eta_{n+1}(t) \leq D_{T} \int_{0}^{t} \eta_{n}(s) d s.
\end{equation}
Where $D_T=2K^2(4+T)$, let $C=\eta_{1}(T)$, we can get the following inequality by recursion,
\begin{equation}
\eta_{n}(T) \leq \frac{C\left(T D_{T}\right)^{n-1}}{(n-1) !}.
\end{equation}
Based on Markov inequality: $P(x \geq a) \leq \frac{E(x)}{a} \text{for}\quad x \geq 0, a>0$, we can get,
\begin{equation}
\sum_{n=1}^{\infty} \mathbb{P}\left[\max _{0 \leq s \leq T}\left|z_{s}^{n}-z_{s}^{n-1}\right| \geq 2^{-n}\right]<\infty.
\end{equation} 
According to Borel-Cantelli lemma \ref{lemma:borel}, we can deduce that the limit $\lim _{n \rightarrow \infty} x_{n}$ exists, and the limit process is continuous. Because the above sequence also converges in the sense of $L^2$, we can take the limit in the recursive equation to get,
\begin{equation}
\mathrm{d} z_{t}=f\left(z_{t} ; t ; \Theta_{f}\right) \mathrm{d} t+g\left(z_{t} ; t ; \Theta_{g}\right) \mathrm{d} B_{t}.
\end{equation}
Hence, the existence of the solution is proved.
\section{The proof of Theorem 2}
\label{sec:appendix-bias-var}
\begin{proof}
The generalization error of the quantile function $\mathcal{F}$ with respect to the ground-truth can be computed as:
\begin{small}
\begin{equation}\nonumber
\begin{aligned}
&\operatorname{MSE}(\mathcal{F},y)={\E}_{(x,y)}[(\mathcal{F}-y_i)^{2}]\\
&={\E}_{(x,y)}[(\mathcal{F}-\E_{(x,y)}(\mathcal{F})+\E_{(x,y)}(\mathcal{F})-y)^{2}]\\
&={\E}_{(x,y)}[(\mathcal{F}-\E_{(x,y)}(\mathcal{F})^2+2(\mathcal{F}-\E_{(x,y)}(\mathcal{F}))(\E_{(x,y)}(\mathcal{F})-y)\\
&+(\E_{(x,y)}(\mathcal{F})-y)^{2}]\\
&={\E}_{(x,y)}(\mathcal{F}-\E_{(x,y)}(\mathcal{F}))^2+{\E}_{(x,y)}[2(\mathcal{F}-\E_{(x,y)}(\mathcal{F}))\\&\quad(\E_{(x,y)}(\mathcal{F})-y)]+{\E}_{(x,y)}(\E(\mathcal{F})-y)^{2}\\
&={\E}_{(x,y)}(\mathcal{F}-\E_{(x,y)}(\mathcal{F}))^2+2(\E_{(x,y)}(\mathcal{F})-y)\\
&\quad{\E}_{(x,y)}(\mathcal{F}-\E_{(x,y)}(\mathcal{F}))+{\E}_{(x,y)}(\E_{(x,y)}(\mathcal{F})-y)^{2}\\
&\operatorname{MSE}(\mathcal{F},y)=\operatorname{Var}_{(x,y)}(\mathcal{F})+\operatorname{Bias}_{(x,y)}(\E_{(x,y)}(\mathcal{F}),y)^2
\end{aligned}
\end{equation}
\end{small}
\end{proof}
\section{Two Side Well Calibration}
More practically, we usually a two-sided calibration to evaluate uncertainty. For a prediction interval (PI) $[F_{i}^{-1}(p_1), F_{i}^{-1}(p_2)]$, $\forall p_1 \in [0,1], p_2 \in [p_1, 1]$, there is a similar definition for two-sided calibration:
\begin{equation}
\frac{\sum_{i=1}^{N} \mathbb{I}\left\{F_{i}^{-1}\left(p_{1}\right) \leq y_{i} \leq F_{i}^{-1}\left(p_{2}\right)\right\}}{N} \rightarrow p_{2}-p_{1},
\label{eqn:cali_two}
\end{equation}
holds as $N \rightarrow \infty$. 
\section{Datasets}
\label{sec:appendix-dataset}
\begin{table}[ht]
    \centering
    \begin{tabular}{cccc}
    \hline
    Datasets& L & D & T\\
    \hline
    Metro-traffic & 48204& 9& 1 hour\\
    Pickups & 29102& 11& 1 hour\\
    Stock &40560 &81 &1 minute\\
    Electricity &26304 &321 &1 hour\\
    Solar & 52560 &137 &10 minutes\\
    Traffic flow & 17544 &862 &1 hour\\
    \hline
    \end{tabular}
    \caption{The description of dataset used, where L is length of time-series or data size, D is number of variables, T is time interval among series.}
    \label{tab:label_dataset}
    \vspace{-0.15cm}
\end{table}
\section{Metrics}
\label{sec:appendix-metrics}
\textbf{Prediction precision metrics.}\quad Root Mean Square Error (RMSE) is the standard deviation of the residuals (prediction errors), which is defined as follow:
\begin{equation}
\operatorname{RMSE}=\sqrt{\frac{1}{n} \sum_{i=1}^{n}\left(y_{i}-\hat{y_{i}}\right)^{2}}
\end{equation}
$R^{2}$ is also a common metric to evaluate forecasting performance, which is a real number in [0,1], the predictions are more accurate when $R^2$ is close to 1, which is defined as follow:
\begin{equation}
    \setlength{\abovedisplayskip}{4pt}
    \setlength{\belowdisplayskip}{4pt}
    R^2 =1-\frac{\sum_{i}\left(y_{i}-\hat{y_{i}}\right)^{2}}{\sum_{i}\left(y_{i}-\bar{y}\right)^{2}},
\end{equation}
\textbf{Calibration metrics.}\quad In the previous works, the expectation of coverage probability error (ECPE)~\cite{cui2020calibrated} of prediction intervals (PIs) is the absolute difference between true confidence and empirical coverage probability, which is defined as:
\begin{equation}
    \text{ECPE} = \frac{1}{n}\sum_{j = 1}^n |P_j - \hat{P}_j|.
\end{equation}
Where $P_j$ is the expected confidence (i.e., the confidence level that we expect), and $\hat{P}_j$ is probability that prediction intervals cover the ground truth.

\textbf{Sharpness metrics.}\quad A kind of sharpness is represented as the averaged width of the prediction intervals (EPIW)~\cite{cui2020calibrated}, which is defined as follow:
    \begin{equation}
    \setlength{\abovedisplayskip}{4pt}
    \setlength{\belowdisplayskip}{4pt}
    \text{EPIW} = \frac{1}{n}\sum_{j = 1}^n \hat{Y}_{j{up}} - \hat{Y}_{j{low}},
    \end{equation}
where n is the total number of prediction intervals, $\hat{Y}_{jup}, \hat{Y}_{jlow}$ are the upper and lower bounds of prediction intervals respectively.
\end{document}